\documentclass[letterpaper, 10 pt, conference]{ieeeconf}  

\IEEEoverridecommandlockouts                              

\usepackage[caption=false,font=footnotesize]{subfig}
\usepackage{graphicx} 
\usepackage{gensymb}

\usepackage{algorithm}
\usepackage{algorithmic}
\usepackage{amsmath}

\title{\LARGE \bf
Robot Risk-Awareness by Formal Risk Reasoning and Planning}
\author{Xuesu Xiao$^{1}$, Jan Dufek$^{1}$, and Robin Murphy$^{1}$
\thanks{$^{1}$Xuesu Xiao, Jan Dufek, and Robin Murphy are with the Department of Computer Science and Engineering,
        Texas A\&M University, College Station, TX 77843
        {\tt\small \{xiaoxuesu, dufek, robin.r.murphy\}@tamu.edu}}%
}

\begin{document}

\maketitle
\thispagestyle{empty}
\pagestyle{empty}

\begin{abstract}
This paper proposes a formal robot motion risk reasoning framework and develops a risk-aware path planner that minimizes the proposed risk. 
While robots locomoting in unstructured or confined environments face a variety of risk, existing risk only focuses on collision with obstacles. Such risk is currently only addressed in ad hoc manners. Without a formal definition, ill-supported properties, e.g. additive or Markovian, are simply assumed. Relied on an incomplete and inaccurate representation of risk, risk-aware planners use ad hoc risk functions or chance constraints to minimize risk. The former inevitably has low fidelity when modeling risk, while the latter conservatively generates feasible path within a probability bound.
Using propositional logic and probability theory, the proposed motion risk reasoning framework is formal. Building upon a universe of risk elements of interest, three major risk categories, i.e. locale-, action-, and traverse-dependent, are introduced. A risk-aware planner is also developed to plan minimum risk path based on the newly proposed risk framework. Results of the risk reasoning and planning are validated in physical experiments in real-world unstructured or confined environments. 
With the proposed fundamental risk reasoning framework, safety of robot locomotion could be explicitly reasoned, quantified, and compared. The risk-aware planner finds safe path in terms of the newly proposed risk framework and enables more risk-aware robot behavior in unstructured or confined environments. 

\end{abstract}

\section{INTRODUCTION}
Robots that locomote in unstructured or confined environments usually face motion risk. Therefore, autonomous planning systems must be capable of deciding how to reduce risk. This intelligent behavior firstly requires a fundamental understanding of risk and then a planning paradigm to find motion plans with minimum risk. The usages of unmanned vehicles in situations such as Urban Search And Rescue (USAR), nuclear operations, disaster robotics \cite{murphy2014disaster}, etc., are examples where the execution of motion inherently entails taking risk and therefore motivate a formal risk reasoning framework and corresponding risk-aware planner. 

\begin{figure}[]
\centering
\includegraphics[width=0.4\columnwidth]{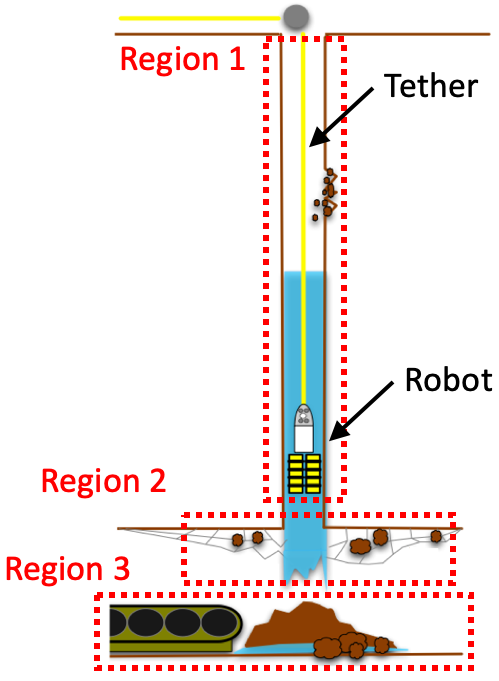}
\caption{Mine Disaster Borehole Entry \cite{murphy2009mobile}}
\label{fig::BE_3regions}
\end{figure}

Fig. \ref{fig::BE_3regions} shows an example of borehole entry from Crandall Canyon Mine (Utah) response in 2007 \cite{murphy2009mobile}. In region 1, the borehole area, small clearance of the borehole may cause the robot getting jammed. Due to the lack of casing of the borehole, falling rocks, drilling foam, water, and debris may damage the robot. The vertically hanging robot might spin and therefore lose controllability and mobility.
In region 2, the transition from the borehole to the mine, the robot may get stuck with the mesh roof during hole exit and reentry. Tether entanglement with mesh roof is another risk. Furthermore, the transition from vertical mobility to operating on mine floor also requires extra effort and induces risk.
Region 3, the inside of the mine, has unstable terrain due to running water and mud, causing the robot getting trapped. The robot also has to traverse soft drill tailings and foam, or even equipment, before reaching the mine floor. Also, while locomoting in region 3, robot tether is still being extended or retracted, interacting with the borehole (region 1) and the mesh roof in the transition into the mine (region 2). Risk of tether entanglement still exists. Due to the variety of existing risk sources in borehole entry, the robot failed at Crandall Canyon Mine in all four runs during the response in 2007 (lowering system failure, blockage in borehole, deteriorated sensing, and entangled and finally broken tether). 

Motivated by the question of how the variety of risk sources contribute to the high failure rate, this paper formally defines robot motion risk using propositional logic and probability theory. In contrast to the conventionally assumed additive or Markovian properties, the formal methods used in the derivation reveals contradicting facts: risk's non-additivity and history dependency. Building upon a comprehensive universe of risk elements (not only obstacles), a variety of adverse effects which exist in unstructured or confined environments are categorized into locale-dependent, action-dependent, and traverse-dependent risk elements. The proposed risk-aware planner is able to take partial history information into account, and plans minimum risk path based on the newly proposed risk reasoning framework (for locale- and action-dependent risk elements). It also provides a paradigm to address risk with deeper history dependency (up to traverse). The risk reasoning and risk-aware planning results are validated through a physical robot flying in a real-world unstructured or confined environment. 

The rest of the paper is organized as follows: Sec. \ref{sec::related_work} provides related work. Sec. \ref{sec::explicit_risk_representation} formally derives the proposed risk reasoning framework. Sec. \ref{sec::approach} presents the risk-aware planner and points out the tradeoff between computation complexity and history dependency depth. Sec. \ref{sec::experiments} presents an integrated demonstration of the risk reasoning and risk-aware planning results on a physical tethered Unmanned Aerial Vehicle (UAV) in a real-world unstructured or confined environment. Sec. \ref{sec::conclusion} concludes the paper. 

\section{RELATED WORK}
\label{sec::related_work}
This section reviews current approaches to reason about risk and existing risk-aware planners. 

\subsection{Risk Reasoning}
Although risk has been addressed in the literature, a formal definition of risk is still unclear. Risk was referred to as some numerical measure of the severity/negativity related with certain aspects of motion. Risk associated with locomoting in unstructured of confined environments was explicitly represented as ad hoc risk functions of robot state. 

\cite{soltani2004fuzzy} represented risk in the workspace by two layers: hazard data and visibility layer. The risk related with each layer was a function of the particular state. Weighted sum was used to combine the effect from both risk layers. To the author's best knowledge, this is the only work that considered more than one risk sources. \cite{de2011minimum} associated UAV flight risk at a certain location with this location's ground orography. \cite{zabarankin2002optimal, gu2006comprehensive} adopted a similar approach and also assumed risk to be a function of location only. Even with data-driven approaches, researchers estimated potential risk of a certain state based on historical record, including ocean Automated Identification System (AIS) \cite{pereira2011toward, pereira2013risk} and traffic data \cite{krumm2017risk}. 

Besides the lack of formal definition, all the risk considered in the literature was related with obstacles of various form. However, moving robots are also exposed to other kinds of risk, such as risk from turning, terrain interaction, battery condition, etc. Furthermore, the assumption that the risk a robot faces at a certain state/location is only a function of that state/location is ill-founded. So is the additivity of risk. Existing work assumed the risk a robot faces when executing a path is simply addition of the risk from all states. 

\subsection{Risk-aware Planning}
In search of a feasible path plan, risk-awareness was mostly modeled as (chance) constraints. \cite{luders2010chance, luders2013robust} proposed chance-constrained rapidly-exploring random tree (CC-RRT) approaches, which used chance constraints to guarantee probabilistic feasibility at each time step and over entire trajectory. Another framework is (PO)MDP. As standard MDP inherently contains reward but not risk, researchers have looked into representing risk as negative reward (penalty) \cite{pereira2013risk} or constraints (C-POMDP) with unit cost for constraint violation \cite{undurti2010online, undurtifunction, undurti2011decentralized}. Going beyond unit cost, CC-POMDP was proposed by \cite{santana2016rao}, which was based on a bound on the probability (chance) of some event happening during policy execution. Robust Model Predictive Control (RMPC) is another alternative, with an emphasis on risk allocation, i.e., to allocate more risk for more rewarding actions \cite{ono2008efficient, vitus2011feedback}. 

Existing constraint violation which models risk took the form of intersection between path and obstacles. A feasible plan was a path with constraint violation probability bounded by a certain manually assigned threshold, not with minimum risk in an absolute sense. Furthermore, the temporal or spatial (multiple obstacles) dependencies of constraint violation probability were either assumed to be independent or relaxed using ellipsoidal relaxation technique or Boole's inequality \cite{santana2016rao, luders2013robust, vitus2011feedback, luders2010chance, ono2008efficient}. This neglected the important dependencies on the motion history and the rough approximation introduced significant conservatism. 

\section{RISK REASONING FRAMEWORK}
\label{sec::explicit_risk_representation}
This section formally defines robot motion risk in unstructured or confined environments and derives the reasoning framework using propositional logic and probability theory. It is comprehensive and general for any robots. 

\subsection{Formal Definiton and Explicit Representation}
This work considers motion risk for mobile robots executing a path. Risk in terms of a sequence of motion (path) is formally defined as \emph{the probability of the robot not being able to finish the path}.

The robot workspace is defined based on tessellation of the Cartesian space. Each tessellation is either a viable (e.g. free) or unviable (e.g. occupied) state for the robot to locomote. A feasible path plan $P$ is defined to be an ordered sequence of viable tessellations, called states and denoted as $s_i$:
\begin{center}
$P = \{s_0, s_1, ..., s_n\},~|| s_i - s_{i-1}||_2 \leq r_c, \forall 1 \leq i \leq n$ 
\end{center}
where $r_c$ is the maximum distance between two consecutive states for the path to be feasible. A state on the path is finished by the robot reaching it within an acceptable tolerance and ready to move on to the next state. A state is not finished due to two main reasons: the robot crashes or gets stuck. In order to finish the path of $n$ states, the robot faces $r$ different risk elements, which will possibly cause not finishing the path. Here, three types of events are defined with propositional logic: 
\begin{itemize}
\item $F$ -- the event where the robot finishes path $P$
\item $F_i$ -- the event where the robot finishes state $i$
\item $F_i^k$ -- the event where risk $k$ does not cause a failure at state $i$
\end{itemize}

The reasoning about motion risk is based on three assumptions, which are expressed by propositional logic: 
\begin{enumerate}
	\item Path is finished only when all states are finished: 
	\begin{center}
	$F = F_n \cap F_{n-1} \cap ... \cap F_1 \cap F_0$
	\end{center}
	
	\item A state is finished only when all risk elements do not cause failure: 
	\begin{center}
	$F_i = F_i^1 \cap F_i^2 \cap ... \cap F_i^{r-1} \cap F_i^r$
	\end{center}
	
	\item Finish or fail a state because of one risk element is conditionally independent of finish or fail that state because of any other risk element, given the history leading to the state:  
	\begin{center}
	$(F_i^1 \vert \bigcap \limits_{j=0}^{i-1} F_j) \perp \!\!\! \perp (F_i^2 \vert \bigcap \limits_{j=0}^{i-1} F_j) \perp \!\!\! \perp ... \perp \!\!\! \perp (F_i^r \vert \bigcap \limits_{j=0}^{i-1} F_j)$
	\end{center}
\end{enumerate}

Before reasoning about risk, the probability of the robot \emph{being} able to finish the path $P(F)$ is firstly reasoned: 

\begin{equation}
\begin{split}
P(F) 
&= P(F_n\cap F_{n-1} \cap ... \cap F_0) \\
&= P(F_n \vert F_{n-1} \cap ... \cap F_0) \cdot ... \cdot P(F_1 \vert F_0) \cdot P(F_0)\\
&= \prod_{i=0}^{n} P(F_i \vert \bigcap_{j=0}^{i-1} F_j) \\
& = \prod_{i=0}^{n} P(F_i^1 \cap F_i^2 \cap ... \cap F_i^r \vert \bigcap_{j=0}^{i-1} F_j) \\
& = \prod_{i=0}^{n} P(F_i^1 \vert \bigcap_{j=0}^{i-1} F_j) \cdot ... \cdot P(F_i^r \vert \bigcap_{j=0}^{i-1} F_j) \\
& = \prod_{i=0}^{n} \prod_{k=1}^{r} P(F_i^k\vert \bigcap_{j=0}^{i-1} F_j)
\end{split}
\end{equation}

The first, second, fourth, and fifth equal signs are due to assumption 1, probability chain rule, assumption 2, and assumption 3, respectively. Therefore, the formal risk definition, the probability of \emph{not} being able to finish the path, is the probabilistic complement: 

\begin{equation}
\begin{split}
P(\bar{F}) 
&= 1 - P(F) \\
&= 1 - \prod_{i=0}^{n} \prod_{k=1}^{r} P(F_i^k \vert \bigcap_{j=0}^{i-1} F_j) \\
&= 1 - \prod_{i=0}^{n} \prod_{k=1}^{r} (1-P(\bar{F_i^k} \vert \bigcap_{j=0}^{i-1} F_j))
\end{split}
\label{eqn::pfbar}
\end{equation}

In terms of risk representation, the risk of path $P$ is denoted as $risk(P)$ and is equal to $P(\bar{F})$. $P(\bar{F_i^k} \vert \bigcap\limits_{j=0}^{i-1} F_j)$ means the probability of risk $k$ causes failure at state $i$, given the history of finishing $s_0$ to $s_{i-1}$. It is therefore denoted as the $k$th risk robot faces at state $i$ given that $s_0$ to $s_{i-1}$ were finished: $r_k(\{s_0, s_1, ..., s_i\})$. Writing in risk representation form will yield: 

\begin{equation}
risk(P) = 1 - \prod_{i=0}^{n} \prod_{k=1}^{r} (1-r_k(\{s_0, s_1, ..., s_i\}))
\label{eqn::risk_representation}
\end{equation}

\begin{figure}[]
\centering
	\includegraphics[width = 0.7 \columnwidth]{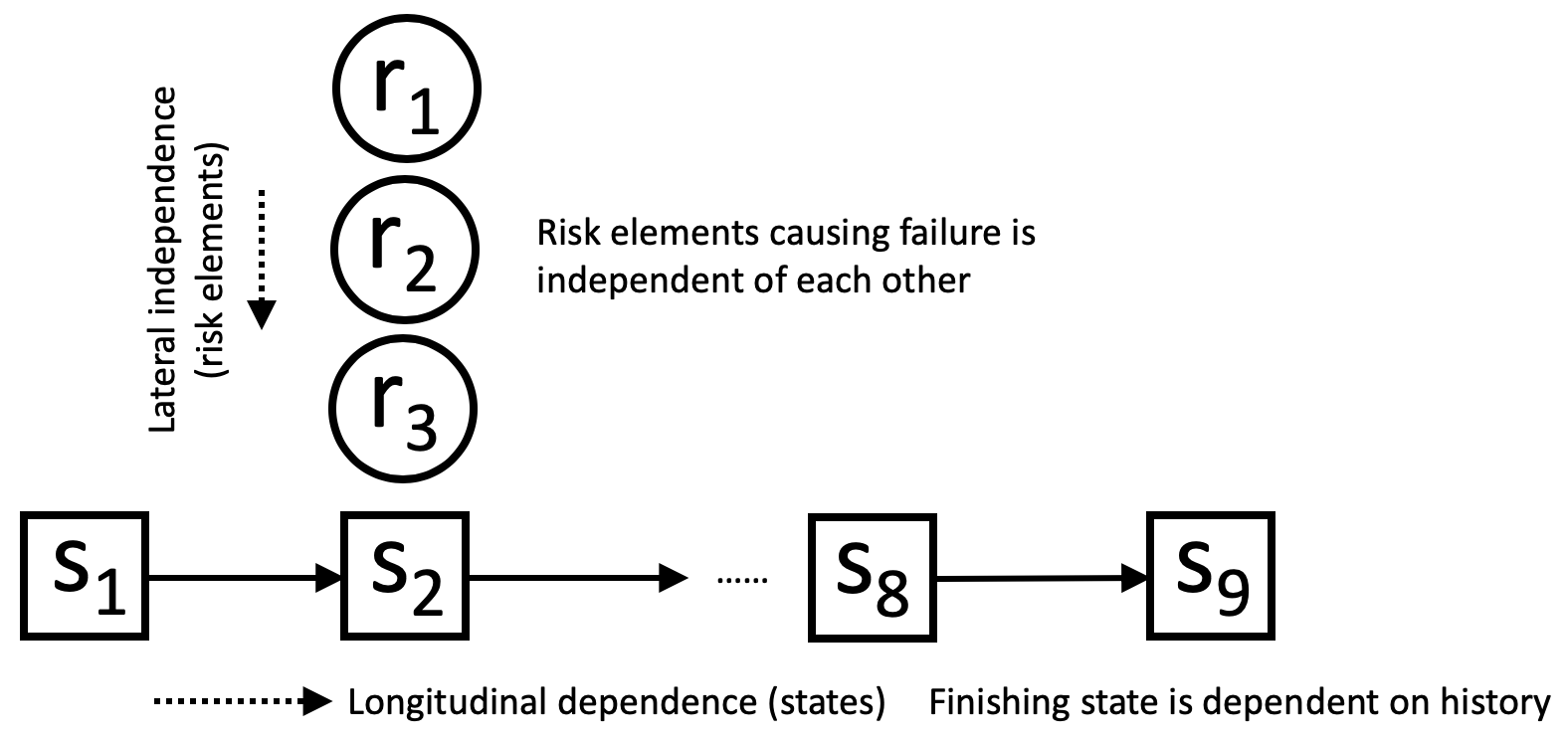}
	\caption{Longitudinal Dependence on History States and Lateral Independence among Risk Elements}
	\label{fig::dependencies}
\end{figure}

This proposed probabilistic motion risk framework does not require the ill-founded additivity assumption for risk. More importantly, the conditional probability in Eqn. \ref{eqn::pfbar} clearly shows the dependency of risk at certain state on the history, not only the state itself. Despite the dependencies in the temporal domain, conditional independence among different risk elements at a certain state given the history is still assumed. Along the direction of the path, risk the robot faces at each individual state is dependent on history (longitudinal dependence), while at each state, risk caused by different risk elements is independent (lateral independence) (Fig. \ref{fig::dependencies}). Note that each individual risk values $r_k(\{s_0, s_1, ..., s_i\})$ could be determined theoretically or empirically. 

\subsection{Risk Categories and Risk Elements}
\begin{figure}[]
\centering
	\includegraphics[width = 1 \columnwidth]{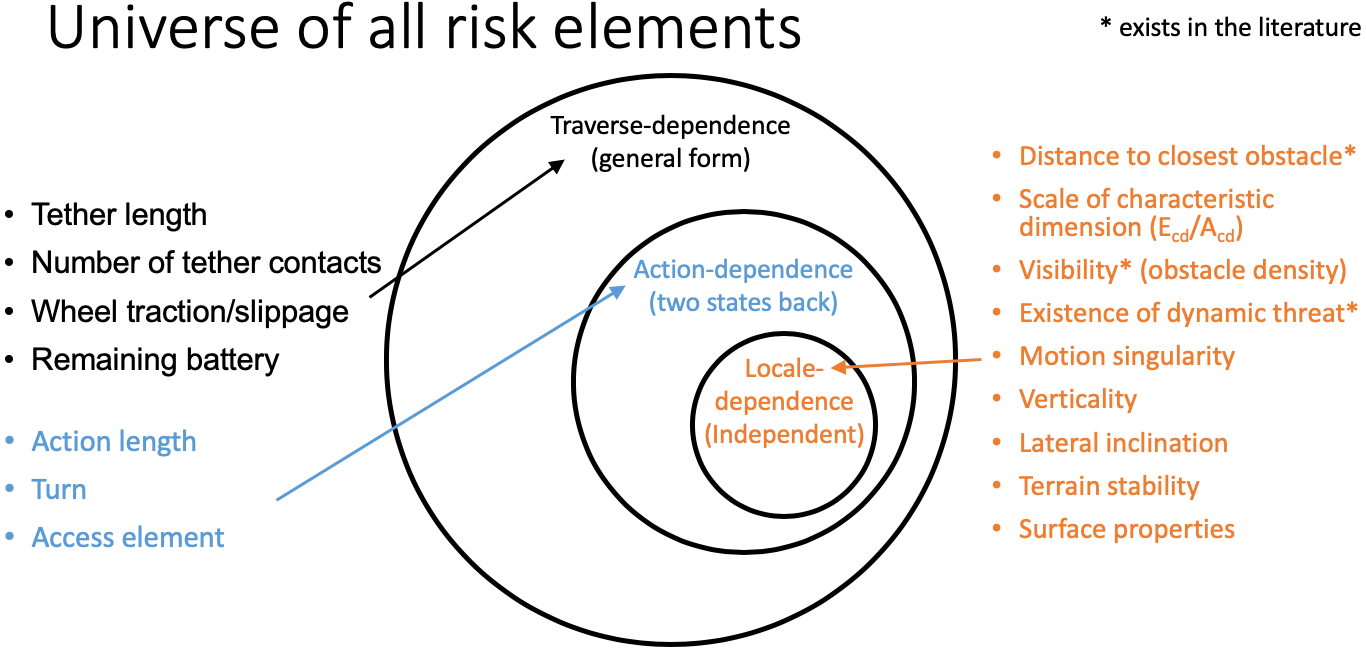}
	\caption{Universe of All Risk Elements}
	\label{fig::universe}
\end{figure}

The formal definition and explicit representation reveal the longitudinal dependence of risk at a certain state on the history. Mathematically speaking, the dependency is on the entire history in general. However, in practice, the dependency of different risk elements may have different depth into the history, e.g. crash to a very close obstacle is only dependent on the closeness of this state to obstacle or crash due to an aggressive turn is only dependent on two states back in the history. In this work, risk elements are divided into three categories: \emph{locale}-dependent, \emph{action}-dependent, and \emph{traverse}-dependent risk elements. Fig. \ref{fig::universe} shows the universe of all risk elements considered, and the categories they belong to. More importantly, the subset/superset relationship between the three categories are displayed: locale-dependence $\subset$ action-dependence $\subset$ traverse-dependence. This universe of risk elements is not exclusive. 

\subsubsection{Locale-dependent Risk Elements}

This is the most special case in history dependence, since its dependency on history could be entirely relaxed. That is: 

\begin{equation}
P(\bar{F_i^k} \vert \bigcap_{j=0}^{i-1} F_j) = P(\bar{F_i^k})
\end{equation}

The word \emph{locale} connotes the meaning of ``location'', ``position'', or where the robot is currently at. It has similar connotation as the concept of ``state'' in (Cartesian) configuration space, but also emphasizes the relationship with the current proximity of the environment. This category of risk elements has been covered in existing literature under the name of ``location'' or ``state'' and was assumed to be the only type of risk elements. This type of traditional risk elements could be evaluated on the state alone, not depending on history. 

\subsubsection{Action-dependent Risk Elements}
This is a special case of risk's history dependency, between the general traverse-dependence and the most special locale-dependence. The depth of action-dependent risk elements' history dependency is two states back, such that the finishing of the last two states has impact on the risk the robot is facing at the current state: 

\begin{equation}
P(\bar{F_i^k} \vert \bigcap_{j=0}^{i-1} F_j) = P(\bar{F_i^k} \vert F_{i-2} \cap F_{i-1})
\end{equation}

This category of risk elements usually focuses on the transitions between states, including the effort necessary to initiate the transition and the difference between two consecutive transitions. 

\subsubsection{Traverse-dependent Risk Elements}
This is the general form of risk's history dependency, which encompasses both locale-dependent and action-dependent risk elements: 

\begin{equation}
P(\bar{F_i^k} \vert \bigcap_{j=0}^{i-1} F_j) = P(\bar{F_i^k} \vert F_{i-1} \cap F_{i-2} \cap ... \cap F_1 \cap F_0) 
\end{equation}

The general form has a full depth of history dependency and looks back to the whole traverse from start leading to the current state. Finishing of all the history states has impact on the finishing of the current state.

\section{RISK-AWARE PLANNING}
\label{sec::approach}
This section develops a risk-aware planner that plans with the newly proposed risk reasoning framework. It guarantees optimality for locale- and action-dependent risk elements. Tradeoff between computation complexity and history dependency depth is discussed. 

\subsection{Impact of Non-additivity and History-Dependency}
The risk-aware planner needs to find minimum risk path between a start and a goal location. Traditional risk-aware planners assumed additive and Markovian properties, therefore their cost function is simply:

\begin{equation}
risk(P) = \sum \limits_{i=0}^{i=1}r(s_i)
\end{equation}

Nonetheless, the cost function based on the proposed risk framework is Eqn. \ref{eqn::risk_representation}. 

This risk representation has neither additivity nor state-dependency, and therefore does not have substructure optimality. The risk robot faces at state $i$ is not well-defined on $s_i$, but can take different values depending on the traverse taken $\{s_0, s_1, ..., s_{i-1}, s_i \}$. 



\begin{figure}[]
\centering
\subfloat[Dynamical]{\includegraphics[width=0.5\columnwidth]{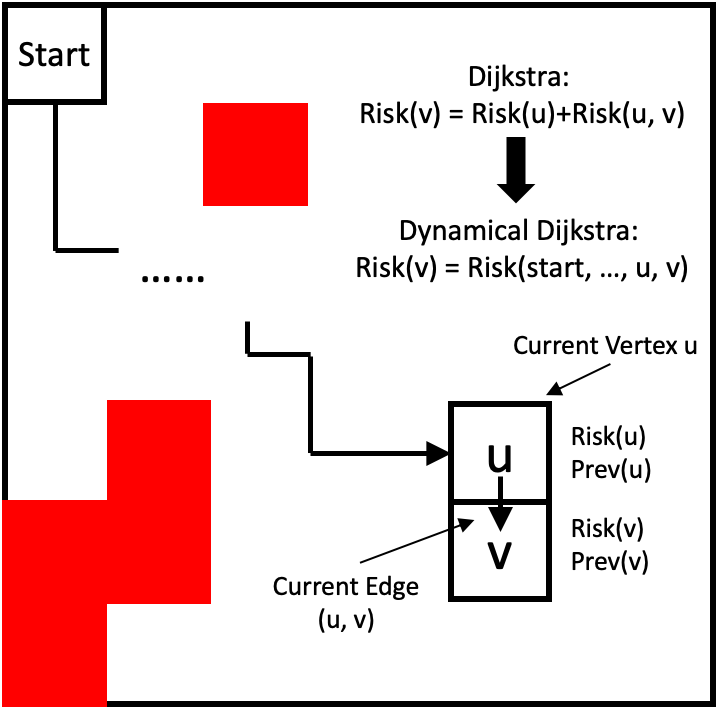}%
\label{fig::dynamical}}
\subfloat[Directional]{\includegraphics[width=0.5\columnwidth]{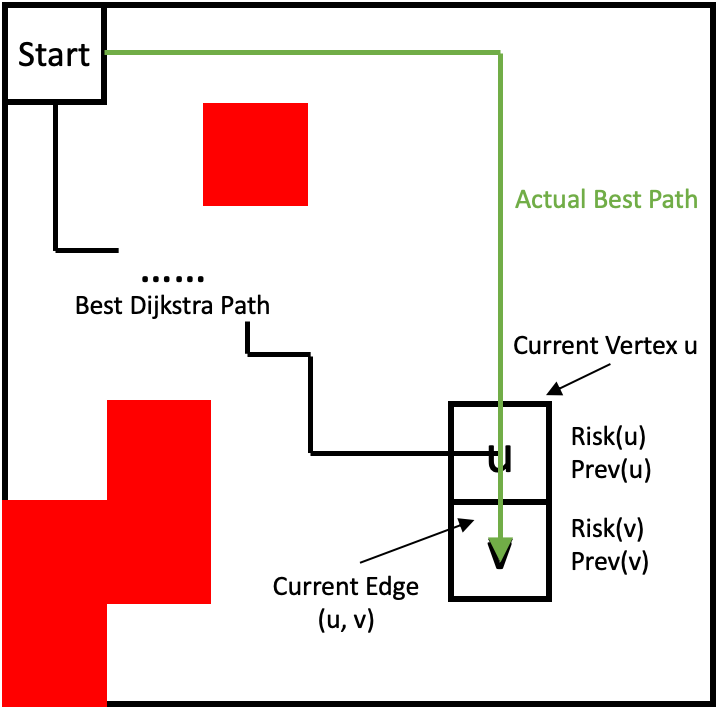}%
\label{fig::directional}}
\caption{Dynamical and Directional Risk Evaluation: Path risk needs to be evaluated dynamically and non-optimal substructure requires minimum risk path to be directional. }
\label{fig::DDD}
\end{figure}

In terms of the impact of those differences on the planner, an intuitive visual example is shown in Fig. \ref{fig::dynamical}: when traditional approaches expand from vertex $u$ to vertex $v$, the risk of the path from start to $v$ is simply the sum of the risk of the subpath from start to $u$ and the risk at $v$. The risk at $v$ is simply well defined on vertex $v$. However, using the proposed risk representation (Eqn. \ref{eqn::risk_representation}), the risk of vertex (state) $v$ is not well defined by only looking at $v$ alone. The dependency on the history requires the risk at $v$ to be evaluated based on the entire traverse (start, ..., $u$, $v$). 

Due to the dynamically changing state risk dependent on history, the problem loses optimal substructure, therefore scenarios such as the one shown in Fig. \ref{fig::directional} may occur: traditional approaches base on substructure optimality, so if the optimal path to $v$ passes through $u$, the subpath to $u$ is also guaranteed to be optimal to $u$. However, due to the risk representation in Eqn. \ref{eqn::risk_representation}, this may not be necessarily true. Given the green path is the best path to $v$, the subpath of the green path to $u$ is not the best path to $u$. The black path actually is. The optimal path to $v$ is from a different direction to $u$ as the optimal path to $u$. 

Therefore the planning problem at hand has at least two issues, preventing from the usage of traditional search algorithms: the risk being dynamical (dependent on history) and directional (lost optimal substructure). Considering the three categories of risk elements, locale-dependency, action-dependency, and traverse-dependency, a new algorithm is designed to optimally address the first two categories of risk elements, by looking back two states into history during planning, at the cost of more computation. Traverse-dependent risk elements, however, cannot be guaranteed to be optimally addressed, since it is the most general form of risk and the look-back has to be into the entire history to guarantee optimality. Only those risk caused by two steps back into the traverse could be handled. 

\subsection{Risk-aware Planner}
The tessellated workspace is defined as a graph $\mathcal{G} = (\mathcal{V}, \mathcal{E})$. Let $\mathcal{V} = \{v_1, v_2, ..., v_n\}$ be the vertex set, and $\mathcal{E} = \{e_1, e_2, ..., e_m\}$ be all the edges connecting the vertices. To accommodate the history dependency of action-dependent risk elements, each vertex is further represented by $v_i = (D_i^{(1)}, D_i^{(2)}, ..., D_i^{(c)})$, where $D_i^{(j)}$ represents the direction from which $v_i$ is reached. They memorize the two-step history information to be used when being expanded in the future. The total number $c$ is the connectivity of $v_i$, as the number of incoming edges reaching $v_i$. For each direction reaching $v_i$, $D_i^{(j)}$ is defined as $D_i^{(j)} = (r_i^{(j)}, PD_i^{(j)})$, where $r_i^{(j)}$ is the risk of reaching $v_i$ from direction $D_i^{(j)}$ starting from start vertex $v_{start}$, and $PD_i^{(j)}$ is the previous direction of reaching the previous vertex, in other words, previous direction of two steps back. All the directions of all vertices $D_i^{(j)}$ compose the superset of all directions $\mathcal{D} = \{D_i^{(j)}|i=1, 2, ..., n\}$ and $j$ is a variable for different vertices depending on how many directions (edges) are leading to the vertex. The algorithm is shown in Alg. \ref{alg::risk-aware}. It finds the minimum-risk directional component in the graph (line 5) and expands the vertex which this directional component belongs to (line 6 - line 15). After expanding all the neighbors, this directional component is marked visited (line 16). When all directional components are visited, the final minimum-risk path to each vertex is selected from its minimum-risk directional components (line 18 - 22). 

\begin{algorithm}[]
 \caption{Risk-Aware Path Planner}
 \begin{algorithmic}[1]
 \renewcommand{\algorithmicrequire}{\textbf{Input:}}
 \renewcommand{\algorithmicensure}{\textbf{Output:}}
 \REQUIRE $\mathcal{G}$, $v_{start}$
 \ENSURE  Risk-Aware paths to all vertices other than $v_{start}$
  \STATE $\forall D_i^{(j)} \in \mathcal{D}$ set $r_i^{(j)} \leftarrow \infty$ and $PD_i^{(j)} \leftarrow NULL$
  \STATE For $v_{start}$, set $r_{start}^{(j)}\leftarrow 0$ in all $D_{start}^{(j)}$
  \STATE Initialize visited set to $\mathcal{R}\leftarrow\{\}$
  \WHILE {$\mathcal{R}\neq\mathcal{D}$}
  	\STATE pick vertex $v_u$ with smallest $r_u^{(i)}$ where $D_u^{(i)}\notin\mathcal{R}$
	\FOR {each edge $(v_u, v_v)\in\mathcal{E}$}
		\STATE $path_u^{(i)}\leftarrow backtrack(D_u^{(i)}$)
		\STATE $path_v(i) \leftarrow path_u^{(i)} \cup \{v_v\}$
		\STATE $path\_risk_v(i)\leftarrow evaluate(path_v(i))$
		\STATE $current\_min\_risk\leftarrow v_v.D_v^{(j)}.r_v^{(j)}$, where $D_v^{(j)}$ corresponds to reaching $v_v$ from $v_u$
		\IF {$path\_risk_v(i)<current\_min\_risk$}
			\STATE $v_v.D_v^{(j)}.r_v^{(j)}\leftarrow path\_risk_v(i)$
			\STATE $v_v.D_v^{(j)}.PD_v^{(j)} \leftarrow D_u^{(i)}$
		\ENDIF
	\ENDFOR
	\STATE $\mathcal{R}\leftarrow \mathcal{R} \cup \{D_u^{(i)}\}$
  \ENDWHILE 
  \FOR {each $v_i \in \mathcal{V}$}
  	\STATE pick $D_i^{(j)}$ with the smallest $r_i^{(j)}$
	\STATE $risk_i \leftarrow r_i^{(j)}$
	\STATE $path_i \leftarrow backtrack(D_i^{(j)})$
  \ENDFOR
 \RETURN all $path_i$ with $risk_i$
 \end{algorithmic}
 \label{alg::risk-aware}
 \end{algorithm}
 
\subsection{Risk Representation and Planning Examples}
\begin{figure}[]
\centering
\subfloat[Path 1 (Planning Result)]{\includegraphics[width=0.5\columnwidth]{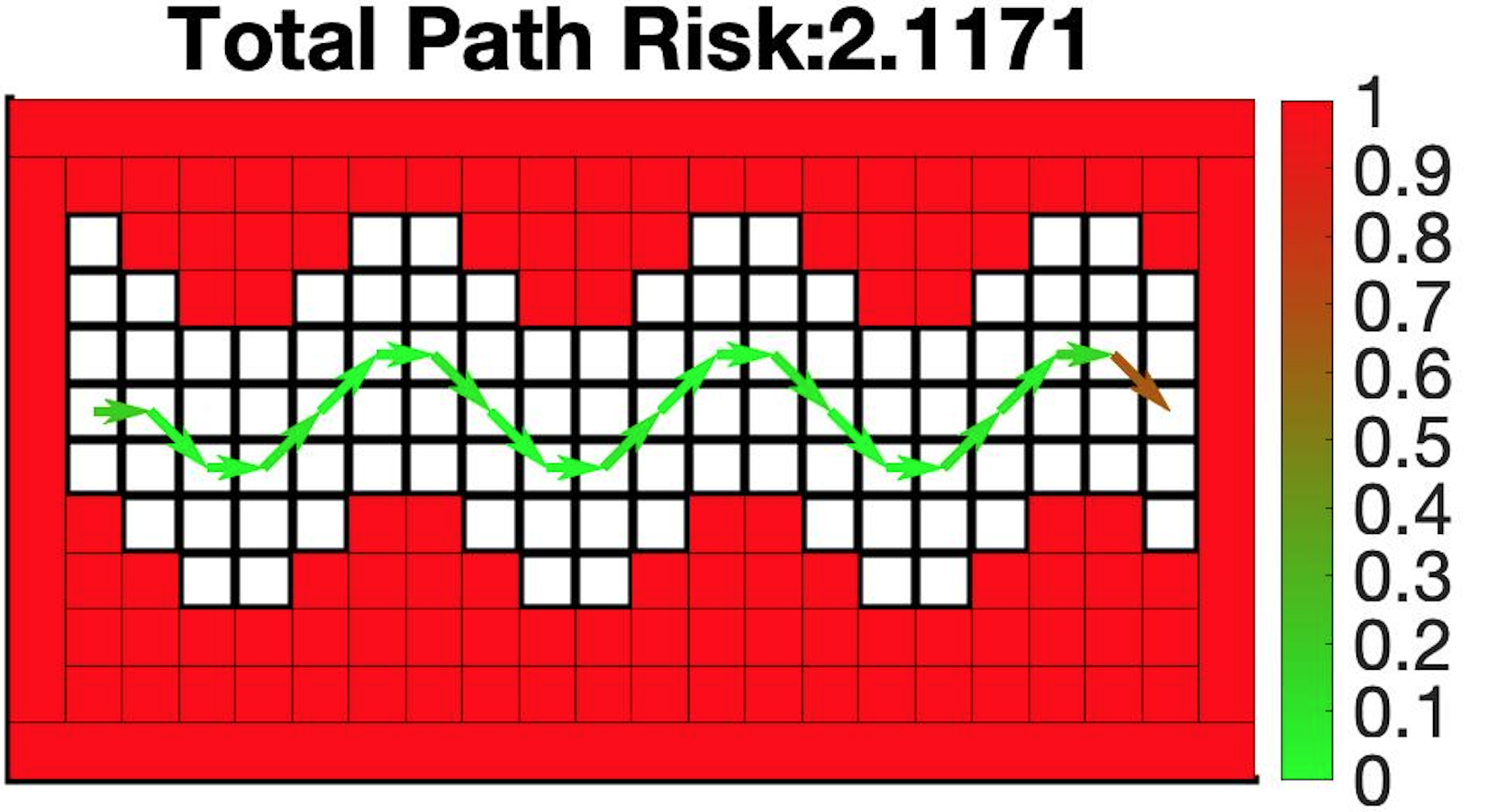}%
\label{fig::path11}}
\hfill
\subfloat[Path 2 (for Comparison)]{\includegraphics[width=0.5\columnwidth]{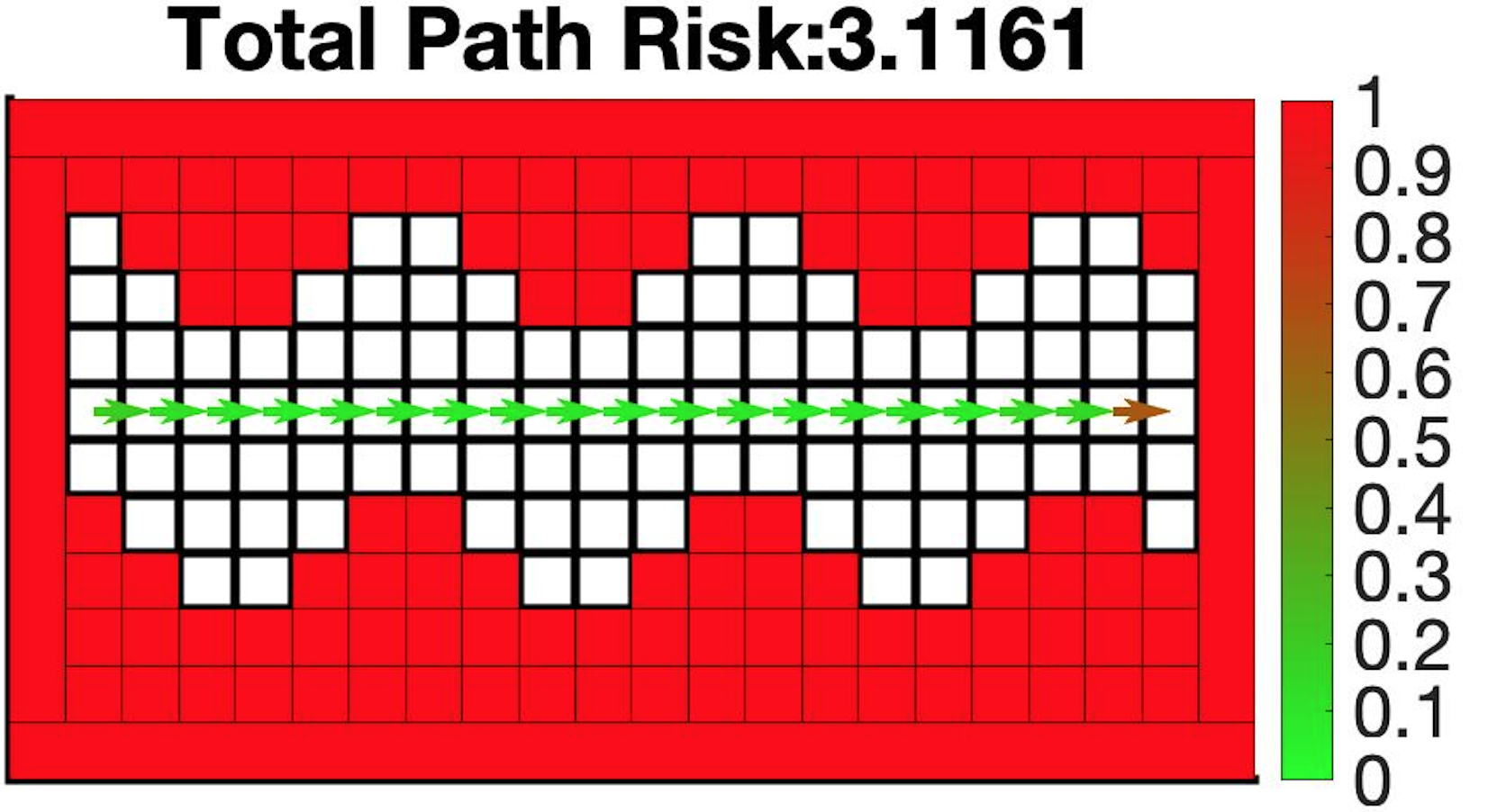}%
\label{fig::path12}}
\caption{Conventional Planner with Additive State-Dependent Risk}
\label{fig::conventional}
\end{figure}

\begin{figure}[]
\centering
\subfloat[Path 1 (for Comparison)]{\includegraphics[width=0.5\columnwidth]{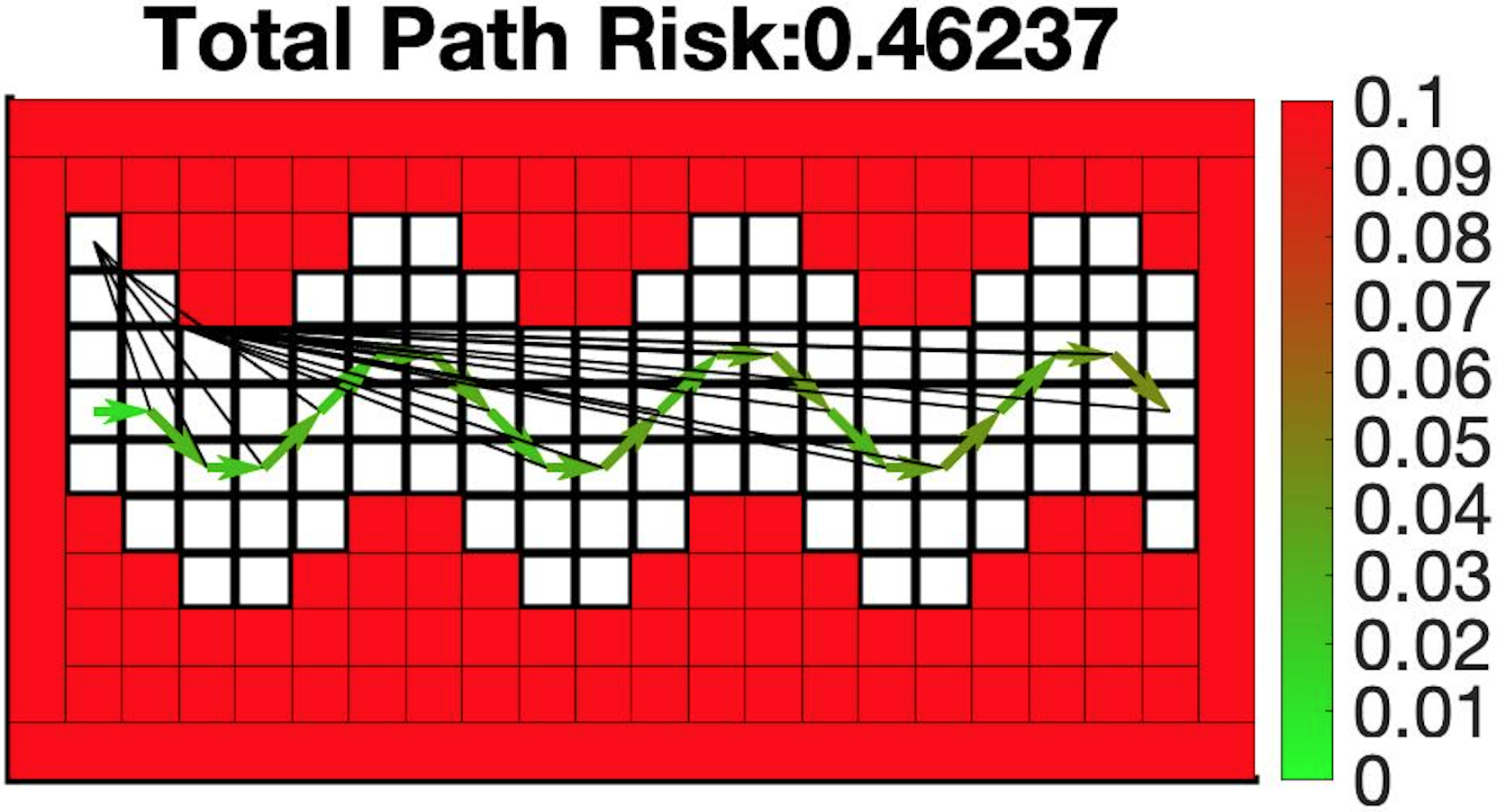}%
\label{fig::path21}}
\hfill
\subfloat[Path 2 (Planning Result)]{\includegraphics[width=0.5\columnwidth]{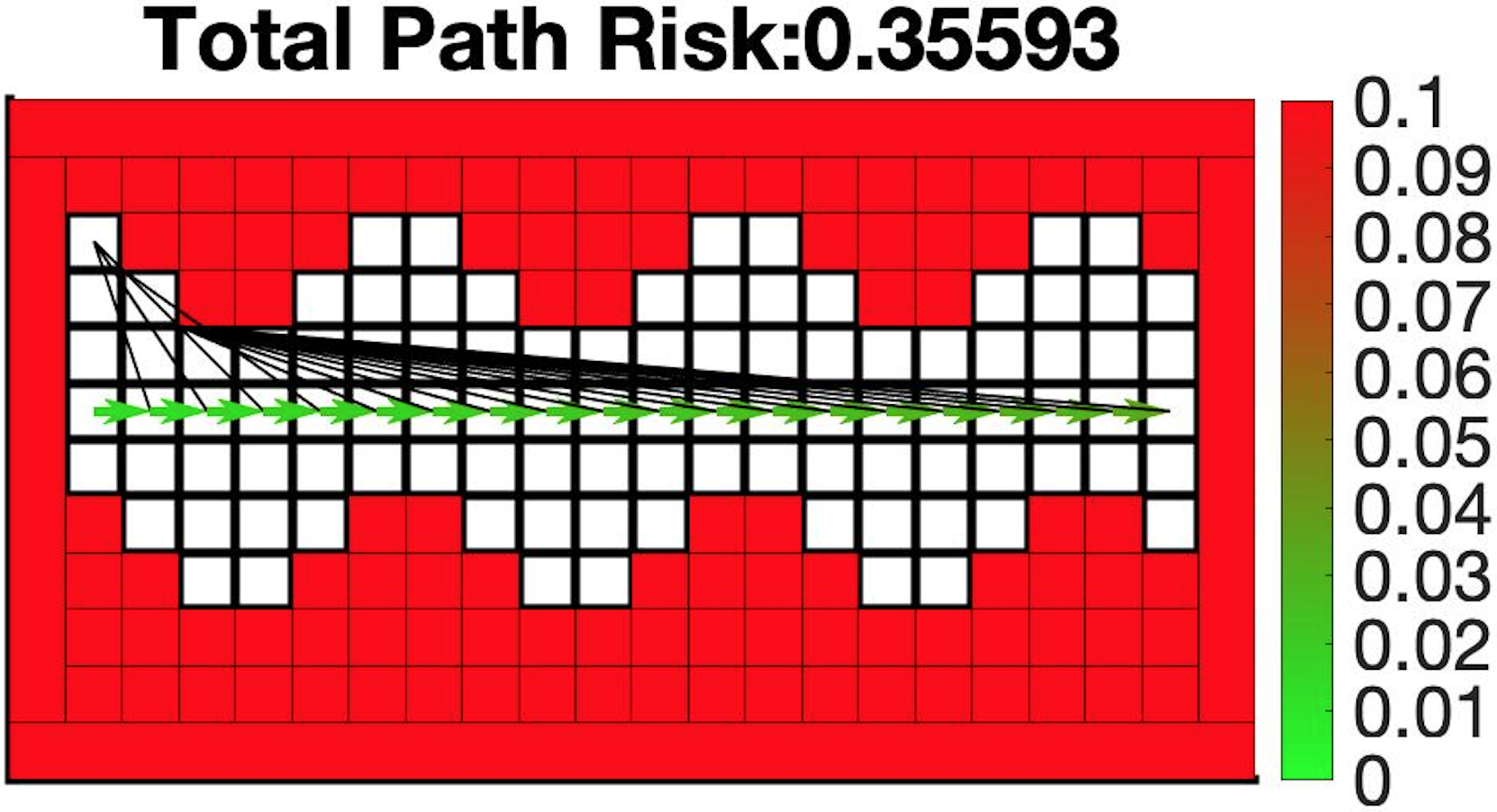}%
\label{fig::path22}}
\caption{Proposed Planner with Probabilistic History-Dependent Risk}
\label{fig::proposed}
\end{figure}

Taking a tethered robot locomoting in obstacle-occupied (shown as red) 2D space as example, results from the proposed risk-aware planner are shown in Fig. \ref{fig::proposed}. As comparison, results of conventional risk-aware planner based on additive state-dependent risk are presented in Fig. \ref{fig::conventional}. The color of the arrows indicates the risk the robot faces at each state and the color map is displayed on the right. The robot starts from the left of the map and the goal is going to the right. Six risk elements are chosen as examples from the three risk categories: distance to closest obstacle and visibility from locale-dependent risk elements, action length and turn from action-dependent risk elements, and tether length and number of tether contacts \cite{xiao2018motion} from traverse-dependent risk elements. Risk caused by each risk element at each state is determined empirically, e.g. a state which is closer to obstacle, requires a sharper turn, or involves more contact points is assigned a higher risk value. 

Fig. \ref{fig::path11} shows the result of conventional risk-aware planner using additive state-dependent risk. Due to the assumption of state-dependency, action length, turn, and tether length, number of contacts cannot be properly addressed by the planner. The only possible risk elements are distance to closest obstacle and visibility, which are evaluated based on state alone. Their risk values at each state are combined using normalization and weighted sum (identical weights for both risk elements in the examples) and summed up along the entire path. Using this approach, the planner will find the path shown in Fig. \ref{fig::path11}, since this is the minimum risk path according to the conventional additive state-dependent risk representation and could be found by traditional search-based algorithms, such as Dijkstra's or A*. The path shown in Fig. \ref{fig::path12}, however, will be neglected, since it is supposed to have a higher risk according to the additive state-dependent risk representation. 

Fig. \ref{fig::path22} shows the result of the proposed risk-aware planner using probabilistic history-dependent risk. All six risk elements from all three risk categories could be properly addressed by the proposed planner, with the optimality of locale-dependent and action-dependent risk elements guaranteed. The risk at each state is now formulated as the probability of the robot not being able to finish the state, displayed in color. The probability of not being able to finish the path, as risk of the path, is computed using Eqn. \ref{eqn::risk_representation}. The two-step look-back in the proposed risk-aware planner makes sure that history dependencies of risk up to actions could be addressed optimally. The traverse-dependent risk elements, however, are only suboptimal, or in other words, optimal up to two states in the history of the traverse, not the entire history to the start. As shown by Fig. \ref{fig::proposed}, the risk aware-planner is willing to sacrifice distance to closest obstacle and visibility (locale-dependent) for shorter action length, less aggressive turn (action-dependent), and shorter tether length, fewer tether contacts (traverse-dependent).

\begin{figure}[]
\centering
	\includegraphics[width = 0.7 \columnwidth]{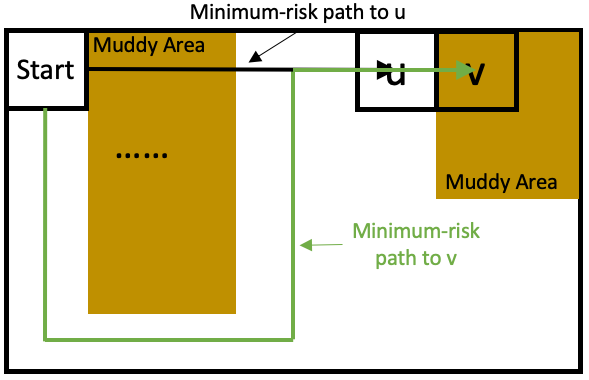}
	\caption{Failure Case due to Traverse-Dependency}
	\label{fig::not_traverse}
\end{figure}

One example of why the proposed risk-aware planner cannot optimally address traverse-dependent risk elements is shown in Fig. \ref{fig::not_traverse}. Take wheel traction/slippage as an example of traverse-dependent risk element and assume the robot has two muddy areas to negotiate with in the workspace: the minimum risk path to $u$ could be the black path, since $u$ is in a clean area and the mud built up on the robot wheels would not cause significant risk at $u$. However, if the robot keeps venturing into $v$, which is another muddy area, the mud built up on the wheels from the first muddy area may cause major risk and the robot has very high probability of getting stuck at $v$. The green path becomes less risky, since the risk associated with the extra length and turns are justified by keeping clean wheels and reliable traction. However, the green path can never been found by the proposed risk-aware planner, since two-step look-back (from $v$ looking back to $u$ and the state left to $u$) cannot cover sufficient depth into history to find the green path. Therefore, for traverse-dependent risk element, only risk caused by the last two steps could be properly addressed, in the similar way as how action-dependent risk element is addressed. 

\subsection{Tradeoff: Computation Complexity vs. Dependency Depth}
\begin{figure}[]
\centering
	\includegraphics[width = 1 \columnwidth]{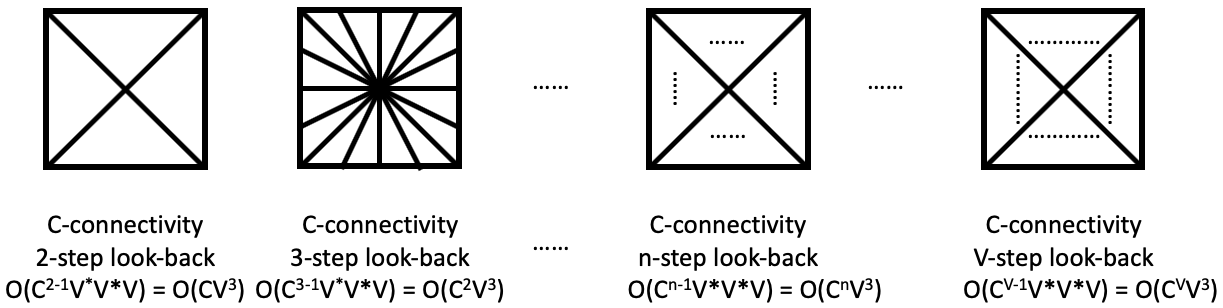}
	\caption{Potential Extension of the Proposed Risk-Aware Planner: Trading more computation for deeper history dependency (Graphical illustrations assume 4-connectivity as example)}
	\label{fig::dependency_computation}
\end{figure}

Although it is shown above that the proposed risk-aware planner is not optimal with respect to general traverse-dependent risk elements, it is possible that look-back into more steps in the history can direct the planner closer to the true optimal path, but at the cost of computation. Fig. \ref{fig::dependency_computation} shows the potential extension of the proposed risk-aware planner in order to be able to address more depth in history dependency. The proposed risk-aware planner looks two-step back into the history and therefore augments every original vertex into $C$ directional components (as the four partitions in the left state of Fig. \ref{fig::dependency_computation}, assuming $C=4$). If three-step look-back is necessary, the original vertex could be augmented into $C^2$ directional components (as the sixteen partitions in the second to left state of Fig. \ref{fig::dependency_computation}, assuming $C=4$). By the same token, an arbitrary number $n$-step look-back requires $C^{n-1}$ directional components. The deepest possible history dependency is $V$ steps, as the longest simple paths have $V$ vertices (here, $V$ is trivially equivalent to $V-1$) and the complexity would be $\mathcal{O}(C^VV^3)$. The complexities of the proposed risk-aware path planner (2-step look-back), potential 3-step look-back planner, general risk-aware planner ($n$-step look-back), and omnipotent risk-aware planner with full history dependency are shown in Fig. \ref{fig::dependency_computation}. The omnipotent risk-aware planner with full history dependency is supposed to guarantee optimality with even traverse-dependent risk elements, but the computation is intractable. 

\section{Integrated Demonstration}
\label{sec::experiments}
This section presents an integrated demonstration using physical robot locomoting in real-world unstructured or confined environment to validate the results of risk reasoning and risk-aware planning. 

\subsection{Robot Mission}
A tethered UAV working as an aerial visual assistant for a tele-operated primary ground robot   \cite{xiao2019autonomous} in a confined staircase is used to validate the results. The marsupial robot team locates at the second level of the staircase. The teleoperation task of the primary ground robot is to insert a sensor between stair railings and drop it into a pool of contamination at the bottom for radioactivity strength reading. This resembles a scenario in Fukushima Daiichi nuclear disaster response. 

\begin{figure}[]
\centering
	\includegraphics[width =1 \columnwidth]{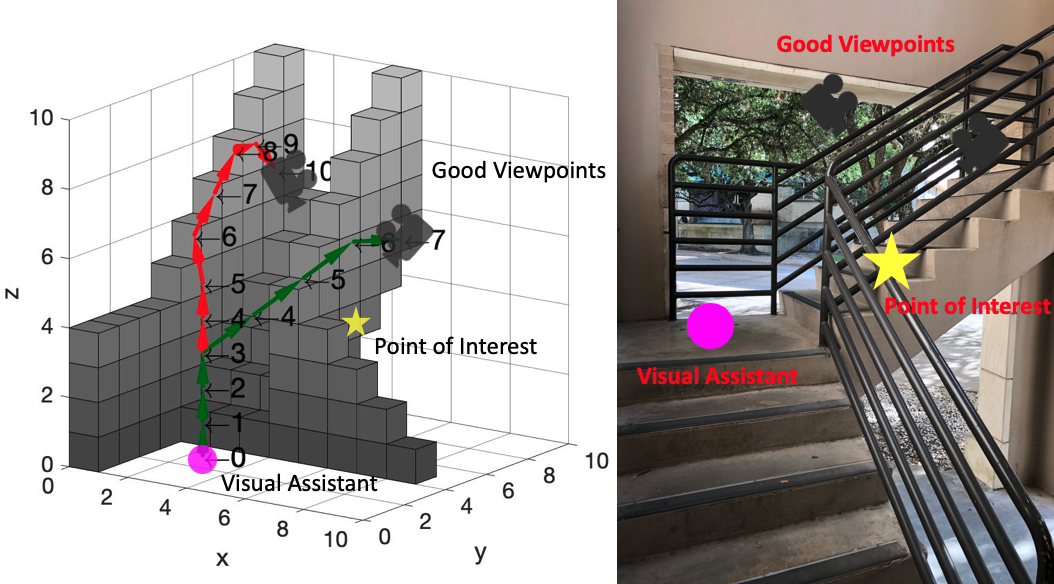}
	\caption{Greyscale voxels represent occupied spaces. Magenta circle shows initial UAV location. Yellow star is the Point of Interest, i.e. the insertion point of the manipulator arm between the railings. The two cameras are good viewpoints.}
	\label{fig::map_and_rewards}
\end{figure}

The map of the environment is given and there exist two good viewpoints to visually assist the sensor insertion task at the point of interest. Based on the proposed risk reasoning framework, the risk-aware planner needs to plan two paths from the UAV initial location to the two good visual assistance viewpoints (Fig. \ref{fig::map_and_rewards}). 

\subsection{Risk Reasoning and Risk-aware Planning Results}
Six different risk elements are considered: distance to closest obstacle (Dist.) and visibility (Vis.) (locale-dependent), action length (A. L.) and turn (action-dependent), tether length (T. L.) and number of tether contacts (Cont. \#) (traverse-dependent). The choice of these six risk elements are due to their relevance to this particular robot platform in this particular unstructured or confined scenario, the practicality or availability of necessary risk information, and the representativeness of the three major risk categories. The minimum risk path to the two good viewpoints are planned, shown in Fig. \ref{fig::map_and_rewards} left. 

The red path aims at one good viewpoint between the two staircase railings. Since a direct path needs to go through spaces confined by staircase railings and walls, the UAV needs to maneuver through those spaces to remain far away from obstacles and high visibility, but at the cost of a longer path and more turns. The risk associated with the red path is evaluated to be 0.714. The detailed risk representation for each state and each individual risk element on the red path is shown in Tab. \ref{tab::red_path_details}. The last column shows the state risk the robot faces at each state. 

\begin{table}[]
\centering
\caption{Detailed Risk Representation for Red Path}
\label{tab::red_path_details}
\begin{tabular}{|c|c|c|c|c|c|c|c|}
\hline
\textbf{Index} & \textbf{Dist.} & \textbf{Vis.} & \textbf{A. L.} & \textbf{Turn} & \textbf{T. L.} & \textbf{Cont. \#} & \textbf{State}\\ \hline
\textbf{0}     & 0.01              & 0.02                & 0                    & 0             & 0.01                 & 0     & 0.04               \\ \hline
\textbf{1}     & 0.01              & 0.02                & 0.04                 & 0             & 0.01                 & 0        & 0.08            \\ \hline
\textbf{2}     & 0.01              & 0.02                & 0.04                 & 0             & 0.01                 & 0       & 0.08             \\ \hline
\textbf{3}     & 0.01              & 0.02                & 0.04                 & 0             & 0.01                 & 0         & 0.08           \\ \hline
\textbf{4}     & 0.01              & 0.01                & 0.04                 & 0             & 0.02                 & 0          & 0.08          \\ \hline
\textbf{5}     & 0.01              & 0.01                & 0.04                 & 0             & 0.02                 & 0      &0.08              \\ \hline
\textbf{6}     & 0.01              & 0.01                & 0.06                 & 0.05          & 0.02                 & 0        & 0.14            \\ \hline
\textbf{7}     & 0.01              & 0.01                & 0.05                 & 0.05          & 0.03                 & 0      & 0.14              \\ \hline
\textbf{8}     & 0.01              & 0.01                & 0.05                 & 0             & 0.04                 & 0      & 0.11              \\ \hline
\textbf{9}     & 0.02              & 0.02                & 0.04                 & 0.05          & 0.04                 & 0         & 0.16           \\ \hline
\textbf{10}    & 0.03              & 0.04                & 0.05                 & 0.05          & 0.04                 & 0     & 0.19               \\ \hline
\end{tabular}
\end{table}

\begin{table}[]
\centering
\caption{Detailed Risk Representation for Green Path}
\label{tab::green_path_details}
\begin{tabular}{|c|c|c|c|c|c|c|c|}
\hline
\textbf{Index} & \textbf{Dist.} & \textbf{Vis.} & \textbf{A. L.} & \textbf{Turn} & \textbf{T. L.} & \textbf{Cont. \#} & \textbf{State}\\ \hline
\textbf{0}     & 0.01              & 0.02                & 0                      & 0             & 0.01                   & 0        & 0.04               \\ \hline
\textbf{1}     & 0.01              & 0.02                & 0.04                   & 0             & 0.01                   & 0       & 0.08               \\ \hline
\textbf{2}     & 0.01              & 0.02                & 0.04                   & 0             & 0.01                   & 0    & 0.08                   \\ \hline
\textbf{3}     & 0.01              & 0.02                & 0.04                   & 0             & 0.01                   & 0    & 0.08                   \\ \hline
\textbf{4}     & 0.03              & 0.01                & 0.06                   & 0.05          & 0.02                   & 0  & 0.16                     \\ \hline
\textbf{5}     & 0.04              & 0.01                & 0.06                   & 0             & 0.02                   & 0  & 0.12                     \\ \hline
\textbf{6}     & 0.02              & 0.01                & 0.06                   & 0             & 0.03                   & 0   & 0.12                    \\ \hline
\textbf{7}     & 0.01              & 0                   & 0.05                   & 0.05          & 0.03                   & 0   & 0.13                    \\ \hline
\end{tabular}
\end{table}

The green path aims at the other viewpoint in the wide open space in the middle of the staircase. Going there straight from the initial location needs to closely pass by the top of the railings. The planner chooses to make a slight detour to enlarge the clearance. However, maximizing distance and visibility has longer path and more turns as cost, so the planner chooses a compromise in between, shown as the 45\degree~middle segment on the green path: the UAV does not fully sacrifice path length and twistiness for clearance, so it cuts through the free space with a straighter path and slightly (not completely) avoids the obstacles. The risk associated with the green path is evaluated to be 0.575 (computed from Tab. \ref{tab::green_path_details}). No contact points are formed in either cases. 

It is worth to note that using the traditional state-dependent only risk representation, the red path has a lower additive risk, because it maintains a relatively low state-dependent risk at most of the states on the path. The green path, however, would have higher risk, due to the compromise of locale-dependent (distance and visibility) for action-dependent risk elements (action length and turns). Although overall the compromise reduces path risk, it cannot be reflected by the traditional state-dependent risk representation. 

\subsection{Experimental Results and Discussions}
Ten experimental trials each are conducted autonomously for the two paths. Path execution is manually terminated when the tethered UAV is unable to finish a state (collision or stuck). Tab. \ref{tab::trials_20} shows the results of the 20 trials, either success or failure. The reasons of failure are specified. 

\begin{table}[]
\centering
\caption{Experimental Trials and Success/Failure}
\label{tab::trials_20}
\begin{tabular}{|c|c|c|}
\hline
Trial \#     & Red Path              & Green Path                         \\ \hline
1            & Collision w. railings & Collision w. wall                  \\ \hline
2            & Oscillation           & \textbf{Success}                   \\ \hline
3            & \textbf{Success}      & Oscillation                        \\ \hline
4            & Collision w. wall     & \textbf{Success}                   \\ \hline
5            & Oscillation      & Collision w. railings              \\ \hline
6            & Oscillation           & \textbf{Success}                   \\ \hline
7            & \textbf{Success}               & Collision w. wall                  \\ \hline
8            & Collision w. wall     & Contacts formed, localization lost \\ \hline
9            & Collision w. railings & \textbf{Success}                   \\ \hline
10           & Oscillation           & Contacts formed, localization lost \\ \hline
Fail. Rate & 0.8                   & 0.6                                \\ \hline
\end{tabular}
\end{table}

For the red path, only two out of the ten experimental trials are successful. The other eight trials fail due to different reasons: trial 1, 4, 8, and 9 fail because collision with railings or wall. Another reason for failure is oscillation. This happens primarily when the UAV is maneuvering to maintain a high obstacle clearance. The turning and long path have the potential of inducing extra turbulence in the confined staircase, therefore the rotorcraft can no longer maintain stability. Oscillation leads to collision or not being able to reach a certain state. The 80\% failure rate is close to the 0.714 risk. Green path execution achieves 60\% failure rate, which is also close to the 0.575 risk value. While in trial 1, 5, and 7 the UAV collides with the obstacles and it starts oscillate in trial 3, another important failure reason is the accidentally formed tether contact due to the closeness to the railings. But overall speaking, the relatively open space in the center of the staircase and the straightness and shortness contribute to a less risky path. Although the risk value caused by each individual risk element is only an empirical estimation, six and eight failures out of ten trials are sufficiently close to the 0.575 and 0.714 risk value, respectively. The proposed theoretical risk reasoning framework matches closely with the real-world failure rate. 

\begin{figure}[]
\centering
\subfloat[Red Path Failure Locations]{\includegraphics[width=0.5\columnwidth]{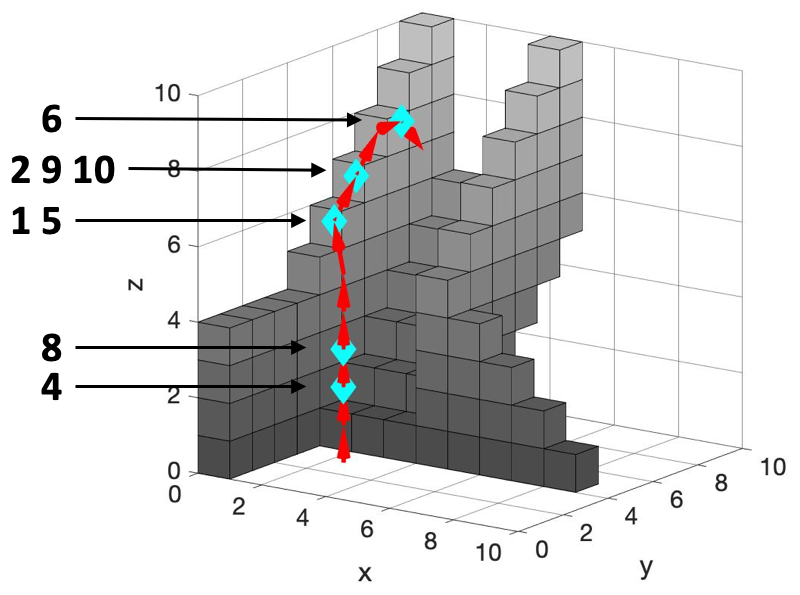}%
\label{fig::red_failures}}
\subfloat[Green Path Failure Locations]{\includegraphics[width=0.5\columnwidth]{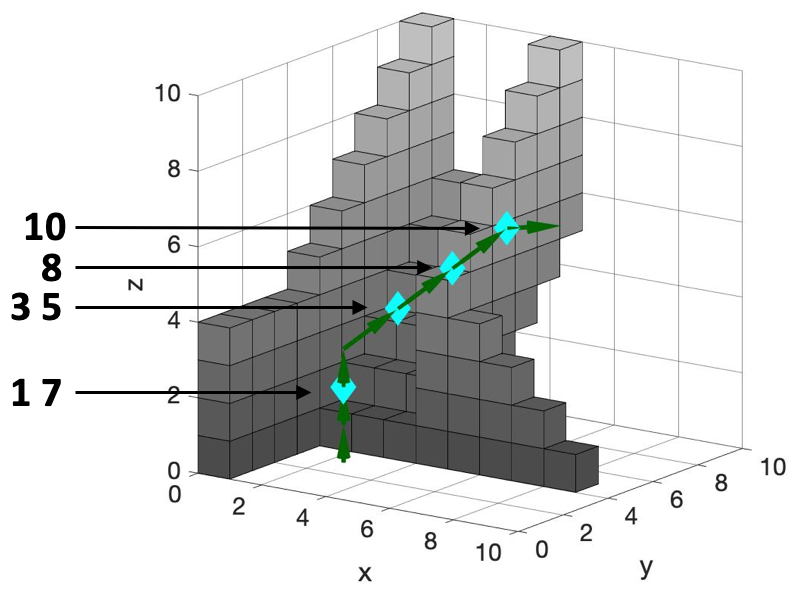}%
\label{fig::green_failures}}
\caption{Failure Locations on Both Paths: The numbers correspond to the trial number in Tab. \ref{tab::trials_20}, indicating this particular trial is terminated at the state denoted by cyan diamonds.  }
\label{fig::failure_locations}
\end{figure}

Fig. \ref{fig::failure_locations} shows the locations of failure. The numbers on the left correspond to the failure trial numbers in Tab. \ref{tab::trials_20}. Some failure locations only have one failure trial, while others may have multiple. Most failure locations for both cases are in the top part of the path, due to either complex trajectory shape (longer path and more turns for red path) or closeness to obstacles (collision or tether contact with obstacles for green path). It roughly matches with the state risk values in the last column of Tab. \ref{tab::red_path_details} and Tab. \ref{tab::green_path_details}: the high state risk values are correlated with more failure cases at that particular state in the physical experiments. Inspecting the failure reasons (Tab. \ref{tab::trials_20}) and failure locations (Fig. \ref{fig::failure_locations}), it could be seen that for red path most failures are caused by action-dependent risk elements while the effect of locale-dependent risk elements is minimized. But for green path, due to the sacrifice of locale-dependent risk elements for shorter path length and fewer turns (action-dependent risk elements), obstacles near states cause more possibility of failure to finish the path. 

\section{CONCLUSION}
\label{sec::conclusion}
This paper proposes a robot motion risk reasoning framework using propositional logic and probability theory. Risk is formally defined as \emph{the probability of the robot not being able to finish the path}. The use of formal methods reveals risk's non-additivity and history-dependency, which was assumed otherwise in the existing literature. Built upon a comprehensive risk universe, three categories of risk elements are introduced: locale-, action-, and traverse-dependent. Based on the proposed risk reasoning framework, a risk-aware planner is developed to take risk's newly discovered properties into account. It finds minimum risk path for locale- and action-dependent risk elements. But for traverse-dependency, optimality cannot be guaranteed. The trade-off between computation complexity and history dependency depth is discussed. The results of applying the risk reasoning framework and risk-aware planner to a tethered UAV flying in a real-world unstructured or confined environment is compared with the proposed theory, and the actual motion failure rate could be roughly reflected by the risk value. Future work will focus on developing better models to compute risk values for individual risk elements and more efficient approaches to extend history dependency depth. 


\section*{ACKNOWLEDGMENT}
This work is supported by NSF 1637955, NRI: A Collaborative Visual Assistant for Robot Operations in Unstructured or Confined Environments. 

\bibliographystyle{IEEEtran}
\bibliography{IEEEabrv,references}

\end{document}